# End-to-End 3D Hand Pose Estimation from Stereo Cameras

Yuncheng Li[1]
raingomm@gmail.com

Zehao Xue[1]
zehao.xue@snap.com

Yingying Wang[1]
ywang@snap.com

Liuhao Ge[2]
ge0001ao@e.ntu.edu.sg

Zhou Ren[3]
renzhou200622@gmail.com

Jonathan Rodriguez[1]
jon@snap.com

[1] Snap Inc.
Santa Monica, CA, USA

[2] Nanyang Technological University
Singapore

[3] Wormpex AI Research

**Abstract**

This work proposes an end-to-end approach to estimate full 3D hand pose from stereo cameras. Most existing methods of estimating hand pose from stereo cameras apply stereo matching to obtain depth map and use depth-based solution to estimate hand pose. In contrast, we propose to bypass the stereo matching and directly estimate the 3D hand pose from the stereo image pairs. The proposed neural network architecture extends from any keypoint predictor to estimate the sparse disparity of the hand joints. In order to effectively train the model, we propose a large scale synthetic dataset that is composed of stereo image pairs and ground truth 3D hand pose annotations. Experiments show that the proposed approach outperforms the existing methods based on the stereo depth.

## 1 Introduction

3D hand pose tracking has many applications in human computer interaction, video games, sign language recognition, augmented reality, *etc*. Solution based on depth sensors has been proven working very well in various conditions [1, 9, 10, 11, 17, 20, 22, 24, 25, 26, 27, 31]. However, due to the hardware constraints, the depth sensors are power consuming, expensive and un-portable, and thus their adoption is limited. To get around the hardware constraints, some recent works proposed methods based on RGB sensors. However, because of the intrinsic scale/depth ambiguity, the methods based on monocular RGB images [2, 14, 19, 23, 33, 35] can only estimate "pseudo" 3D hand pose, which are root relative and scale normalized and the coordinates in the world space cannot be recovered. To estimate full 3D hand pose and leveraging the portability of the RGB sensors, pioneering works are proposed to estimate 3D hand pose from stereo images [18, 34]. Specifically, Zhang *et al*.

proposed a dedicated stereo matching pipeline to generate high quality depth map for hands and leverage existing depth base solution to estimate 3D hand pose [34]. Also using stereo cameras, Panteleris *et al*. proposed a model-based solution to fit the 3D hand pose directly from stereo pairs, bypassing the need to estimate depth map [18]. Although these methods took advantage of the more portable stereo cameras, they are still impractical for real-time application on embedded devices, due to the large computational complexity of performing stereo matching [34] or model fitting [18]. To improve 3D hand pose tracking for portable embedded devices, in this paper, we propose an end-to-end approach to estimate the full 3D hand pose from stereo cameras.

While the end-to-end deep learning-based approach has become the mainstream for body/hand pose estimation using monocular RGB images, it is non-trivial to extend them to stereo RGB images. There are a few unique challenges to work with stereo image pairs. Firstly, there is no extensive datasets to properly train and evaluate the methods. The dataset proposed in [34] was designed to evaluate depth-based solution, so the variance in the dataset is limited. To tackle this challenge, we propose a large scale synthetic dataset. The synthetic stereo RGB pairs are rendered using two virtual cameras with the same camera parameters as in the public STB dataset. The rendering software enumerates variances in hand poses, hand shapes, lighting conditions and skin colors.

Secondly, unlike monocular images, it is more restrictive to apply geometric transformations to the images, because some geometric transformations (such as rotation) breaks the assumption in the rectified stereo system. However, extensive data augmentation is critical to train well performing neural networks. To tackle this challenge, we propose a two stage training procedure to train the networks sufficiently. Namely, we first train the weights related to 2D estimation with extensive data augmentation, and then train the weights related to 3D estimation with allowed data augmentation.

To the best of our knowledge, this work is the first to estimate full 3D hand pose from stereo cameras in an end-to-end framework, so we focus on establishing the evaluation protocols and tackle the unique challenges in this task. To summarize our contributions, 1) We propose a large scale synthetic dataset to estimate full 3D hand pose from stereo cameras. 2) We propose an efficient and flexible framework to extend the existing 2D pose estimation networks to estimate the sparse disparity on hand joints. 3) We show that our proposed end-to-end approach outperform the methods based on stereo depth.

# 2   Related Work

Depth based 3D hand pose estimation works very well in various conditions. There are three kinds of methods based on depth solution, generative [1, 10, 17, 20, 27], discriminative [9, 11, 26, 51] and hybrid [22, 24, 25]. Recent works using deep learning techniques pushed the frontier even further, such as the iterative feedback [16], cascade spatial attention [52], region ensemble network [5], generative adversarial network (GAN) [28], HandPointNet [3], *etc*.

In recent years, researchers are seeking more portable solutions, primarily based on monocular RGB images, such as the hand pose prior [55], depth map as weak supervision [2], variational methods [23], cycle GAN [14], model fitting approach [19], privileged learning [53], graph CNN [4] *etc*. Unfortunately, monocular RGB images have their intrinsic scale/depth ambiguity, so the methods based on monocular images can only estimate "pseudo" 3D hand pose that are root relative and scale normalized.

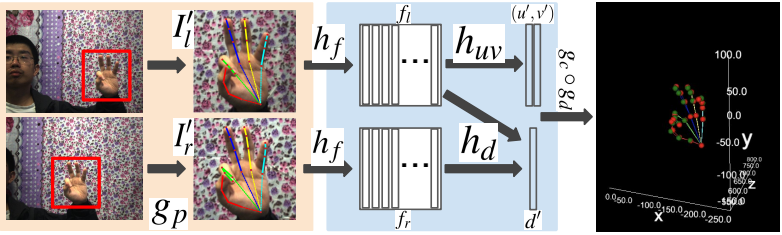

Figure 1: Overview of our method for 3D hand pose estimation from stereo cameras.

Also, to estimate the real 3D hand pose using portable devices, there are methods based on stereo cameras. There are two methods using stereo camera that are most relevant to our work. The first one is the approach proposed by Zhang *et al.*, based on a dedicated stereo matching pipeline that designed specifically for hand pose estimation using manually designed features [34]. The other one is the approach proposed by Panteleris *et al.* [18], based on a model fitting algorithm. Both methods are computationally complex, because of the stereo matching process [34] or the model fitting algorithm [18].

Existing works on **deep learning for stereo images** are also relevant to this work [8, 12, 21]. Most of the existing works used deep learning to improve stereo matching, and use the generated depth for upstream tasks. In this work, we propose to bypass the stereo matching and perform the high level tasks directly.

# 3  Our Method

## 3.1  Formulation

An illustration of the proposed framework to estimate full 3D hand pose from stereo cameras is shown in Figure 1, which can be formulated as the following function,

$$e : (I_l \in \mathbb{R}^{H \times W \times 3}, I_r \in \mathbb{R}^{H \times W \times 3}) \mapsto Y \in \mathbb{R}^{J \times 3}, \quad (1)$$

where $W$ and $H$ denote the width and height of the input images (respectively), $I_l$ and $I_r \in \mathbb{R}^{H \times W \times 3}$ denote the rectified image pairs from the stereo cameras [1], $J$ denote the number of joints on hand (in our experiments, $J = 21$), and $Y$ denote the coordinates of the hand joints in the left camera world coordinate space (without loss of generality, we assume left and right rectified stereo camera system).

It is more intuitive for neural networks to estimate measures that are defined directly on the image space [13], so in our framework, instead of estimating world space $(x, y, z)$, we choose to estimate the 2D location and the disparity that are well defined in the image space. Specifically, in a left/right stereo camera system, the joint coordinates $Y = (x, y, z) \in \mathbb{R}^{J \times 3}$ can be decomposed into the 2D location ($u \in [1, W]^J$, $v \in [1, H]^J$) and the disparity ($d \in [0, W]^J$) by a fixed transformation $g_c : (u, v, d) \mapsto (x, y, z)$. Under the assumption of pinhole camera model, the transformation $Y = g_c(u, v, d)$ is defined as follows,

$$z = f_x \times B / d; \quad x = (u - t_x) / f_x * z; \quad y = (v - t_y) / f_y * z \quad (2)$$

where $f_x, f_y, t_x, t_y$ and $B$ are constants defined by the stereo camera system. Specifically, $f_x$ and $f_y$ denote the focal length expressed in pixel units, $t_x$ and $t_y$ denote the principal point (usually the image center) and $B$ denote the *baseline* parameter (*i.e.*, the distance between the two cameras). Next, we formulate the model to estimate $(u, v, d)$ from the stereo pair.

---

[1]For example, OpenCV has a function named *stereoRectify* to rectify stereo cameras.

As with other pose estimation system based on neural networks, the input images are first cropped and resized so that the target is roughly in the center before feeding into the neural network. For monocular image input, the image is cropped according to a bounding box $(u_0, v_0, w_0, h_0)$, where $(u_0, v_0)$ and $(w_0, h_0)$ are the top-left corner and the size of the crop bounding box, respectively. Then, the cropped image is resized to the network input size $(W_n, H_n)$. For stereo image pair input, we propose to process the left image in the same way as in the monocular image case, but the right image is cropped with the shifted bounding box $(u_0 - d_0, v_0, w_0, h_0)$, where $d_0$ is the global disparity. The reason of using shifted bounding box to crop the right image is to make sure the hand is inside both of the cropped images. Note that different from the traditional stereo matching algorithm, the preprocessed disparity contain negative values because of the shifted right box. How to obtain the initialization parameters $\Phi_0 = \{u_0, v_0, w_0, h_0, d_0\}$ are explained in the Section 4.1. Let us denote this preprocessing step as $g_p : (I_l, I_r) \mapsto (I'_l, I'_r)$.

A properly trained neural network then takes the preprocessed image pairs $(I'_l, I'_r)$, and generate local estimation of 2d location $(u', v')$ and disparity $d'$. The details of this neural network are explained in Sec. 3.2 and the loss functions to train this network are explained in the Sec. 3.3. Let $g_n$ denote the neural network inference: $g_n : (I'_l, I'_r) \mapsto (u', v', d')$. Because the neural network inputs are preprocessed, the estimation are relative to the preprocessing parameters. The denormalization process $g_d : (u', v', d') \mapsto (u, v, d)$ to recover the global estimation are defined as follows,

$$u = u'/W_n * w_0 + u_0; \quad v = v'/H_n * h_0 + v_0; \quad d = d'/W_n * w_0 + d_0 \tag{3}$$

In summary, the system to perform full 3D hand pose estimation from stereo cameras are decomposed by preprocessing step $g_p$, neural network inference $g_n$, denormalization $g_d$, and inverse camera projection $g_c$, *i.e.*, $e = g_c \circ g_d \circ g_n \circ g_p$. In the next sections, we introduce the neural network to implement $g_n$ and the loss functions to train the neural network.

## 3.2    Network Architecture

The neural network $g_n : (I'_l, I'_r) \mapsto (u', v', d')$ can be extended from any 2D keypoint regressor (*e.g.*, Hourglass [15], CPM [29], Simple [30]) by adding components to estimate the disparity. Any 2D keypoint regressor can be expressed by two components, feature extraction and prediction head, both of which are implemented with basic CNN building blocks. Denote the feature extraction block as $h_f : I' \mapsto f \in \mathbb{R}^{H_f \times W_f \times C_f}$, where $f$ are the extracted feature maps, and $H_f, W_f, C_f$ are the height, width and channel number of the feature maps. Denote the 2D location prediction head as $h_{uv} : f \mapsto (u', v') \in \mathbb{R}^{J \times 2}$.

In the proposed neural network, the 2D coordinates are produced directly from the left image using a 2D keypoint regressor [15], *i.e.*, $(u', v') = h_{uv} \circ h_f(I'_l)$. The disparity is estimated by reusing the previous feature extraction network and using it to extract the features from both images. Specifically, the disparity is estimated using the following equations,

$$f_l = h_f(I'_l); f_r = h_f(I'_r); f_{lr} = Concat(f_l, f_r); d' = h_d(f_{lr}), \tag{4}$$

where *Concat* is the standard concatenation, $f_{lr} \in \mathbb{R}^{H_f \times W_f \times 2C_f}$ is the concatenated feature maps, $h_d : \mathbb{R}^{H_f \times W_f \times 2C_f} \mapsto \mathbb{R}^{J \times 1}$ is the network component to predict disparity $d'$. Note that in order to reuse the network weights and computation, the 2D regressor and the disparity estimation share the same feature extraction component.

Two characteristics of the sparse disparity estimation task guide the design of the disparity head $h_d$. 1) Although the task is to estimate disparity on different joints, the underlining algorithm are the same for all the joints, *i.e.* find the best match on the horizontal line. 2) The component should be translation invariant. In other words, if the input images shift horizontally or vertically, the output should be the same. Based on these observations, we propose to use fully convolutional neural network (*FCN*) to estimate the disparity $d \in \mathbb{R}^{J \times 1}$. Specifically, we first use CNN blocks to generate a single channel disparity map from the feature maps, denoting this process as $h_D : f_{lr} \mapsto D \in \mathbb{R}^{H_d \times W_d}$. Size of the disparity map, $H_d$ and $W_d$, are smaller than the network input size to save computation, *i.e.*, $H_n = sH_d$, $W_n = sW_d$ ($s \geq 2$ is the integer stride factor). Then, given the estimated 2D location $(u', v') \in \mathbb{R}^{J \times 2}$ of the joints, we sample the sparse disparity $d'$ from the disparity map $D$ using bilinear interpolation, denoted as follows,

$$d'_j = \sum_{m=1}^{H_d} \sum_{n=1}^{W_d} D[m,n] G(m, u'_j/s) G(n, v'_j/s), \qquad (5)$$

where $j$ is the index of the joint, $G(p,q) = max(0, 1 - |p - q|)$ are the bilinear interpolation coefficients, $D[m,n]$ is the value of $D$ at $m^{th}$ row and $n^{th}$ column.

## 3.3 Loss functions

There are two prediction heads in the proposed network $g_n$, *i.e.*, $h_{uv}$ for 2D location and $h_d$ for the disparity. For $h_{uv}$, we adopt the commonly used heatmap loss, denoted as $L_{uv}$. For $h_d$, the network output is a disparity map $D \in \mathbb{R}^{H_d \times W_d}$, but the available supervision is the 2D location and sparse disparity of the joints, *i.e.*, $(u^{gt}, v^{gt}, d^{gt}) \in \mathbb{R}^{J \times 3}$ [2]. In order to supervise the disparity map $D$ with the ground truth, and to be consistent with the inference algorithm in Eqn. (5), we propose a loss function based on the 2D heatmap. Specifically, as in the heatmap loss, we first build normalized 2D heatmaps for each joint,

$$H_j[m,n] = A_j exp \left( - \left( \frac{(n - u_j^{gt}/s)^2}{2\sigma^2} + \frac{(m - v_j^{gt}/s)^2}{2\sigma^2} \right) \right), \qquad (6)$$

where $\sigma$ is the standard deviation of the heatmap (set to 3 in the experiments), and $A_j$ is the normalization scalar to make $H_j$ sum up to one. The heatmaps $H_j[m,n]$ can be viewed as the probability of the joint $j$ appears at $(m,n)$. Therefore, we propose to use the following loss function to supervise $h_d$,

$$L_d = (1/J) \sum_j \| d_j^{gt} - \sum_{m,n} H_j[m,n] D[m,n] \|_\delta, \qquad (7)$$

where the second term $\sum_{m,n} H_j[m,n] D[m,n]$ is the expectation of the disparity of the $j^{th}$ joint under the probability $H_j[m,n]$, and $\| * \|_\delta$ is the Huber loss with threshold $\delta$ (set to 1 at the experiments). The loss function $L_d$ enforce the network to focus on the sparse disparity on the joints, rather than the dense disparity on the full image. This is one of the key reasons why this approach is more efficient than the methods based on dense stereo matching.

---

[2] $(u^{gt}, v^{gt}, d^{gt})$ are also preprocessed by the preprocessing step $g_p$, but we omit the prime superscript for clarity

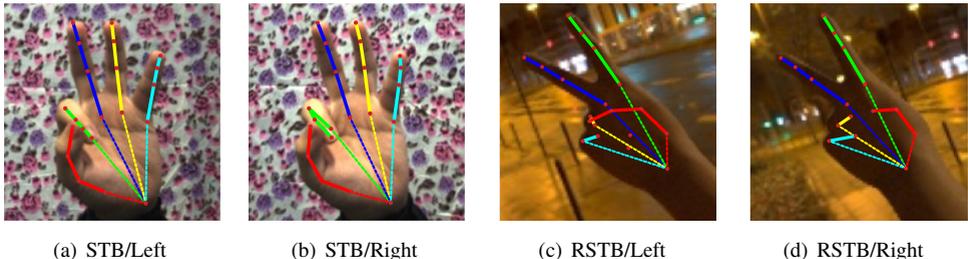

|     (a) STB/Left     |     (b) STB/Right     |     (c) RSTB/Left     |     (d) RSTB/Right     |

Figure 2: Dataset examples. Left/Right denotes the left/right views in the stereo pair. STB/RSTB are the public real and proposed synthetic dataset, respectively. Groundtruth 3D annotation is projected to the two views and visualized with the colored lines.

In order to train the neural network properly, we propose to optimize $L_{uv}$ and $L_d$ separately, rather than train all weights together with $L_{uv} + L_d$. Specifically, the training process is split into two stages. The first stage optimizes $L_{uv}$, i.e., $h_f^*, h_{uv}^* = argmin_{h_f, h_{uv}}(L_{uv})$, and the second stage optimize $L_d$, i.e., $h_d^* = argmin_{h_d}(L_d)$, while fixing $h_f^*$. There are three factors driving this choice of training the 2D location regressor first, 1) the 2D regressor is well developed and understood by the community, so there are more references and resources; 2) there are more annotated dataset for the 2D regressor, and more dataset can generally improve feature representation; 3) there are more freedom to do data augmentation for the 2D regressor than for the disparity head, because some data augmentation, such as rotation, will break the assumption of stereo systems.

## 4 Experiments

### 4.1 Datasets, Protocols and Implementations

For 3D hand pose estimation from stereo cameras, there are two publicly available datasets, one from Zhang et al. [34] and the other from Panteleris et al. [18], which are referred as STB and B2RGB-SH, respectively. Because the B2RGB work [13] is a generative model (does not require training), B2RGB-SH only contains a small dataset for evaluation, which is not suitable for our case that requires training. Therefore, we evaluate the proposed system on the STB dataset, and use the same splits as used in Zimmermann et al. [35]. In addition to the STB dataset, we propose a large scale synthetic dataset to account for large variances in gestures, hand shapes and backgrounds. Specifically, we use software to synthesize hands with various hand shapes, gestures, skin colors and light conditions, and then we render the hands to two virtual cameras to simulate the stereo camera setup. For the two virtual cameras, we use the same camera parameters as used by the STB dataset. Because the hand model is synthesized, the groundtruth (2D location and disparity) is readily known. Besides the rendered foreground hands in stereo image pairs, we use a real stereo camera, similar with the camera used by the STB dataset, to capture background videos in the wild that do not contain any hands. Because the real stereo camera and the virtual cameras share the same intrinsic parameters, we are able to simply alpha blend synthesized foreground and in the wild background together to generate a training sample. For simplicity, we do not use more advanced blending than just overlaying the hand region on top of the background. Figure 2(c) and 2(d) show an example in our synthetic dataset. The blending is done on the fly, so that the effective number of training samples is very large. For the synthetic dataset, there are 18000 pairs for training and 2000 pairs for validation. For the in the wild background, there are 13982 pairs for training and 2419 pairs for validation. We denote the synthetic dataset as *Rendered STB (RSTB)*. The RSTB datasets will be released for research use.

|  | 2D/Frame | 2D/Track | 3D/Frame | 3D/Track |
|---|---|---|---|---|
| clockwise rotate by $[-20°, 20°]$ | Yes | Yes | No | No |
| shift $(u_0, v_0)$ by $[-20\%, 20\%]$ | No | Yes | No | No |
| shift $d_0$ by $[-10\%, 10\%]$ | No | No | No | Yes |
| scale $(w_0, h_0)$ by $[-20\%, 20\%]$ | Yes | Yes | Yes | Yes |

Table 1: Data augmentation under different conditions. *2D/3D* denote the two training stages and *Frame/Track* denote the two evaluation protocols.

As mentioned in Sec. 3.1, the neural network inferences rely on a set of initialization parameters $\Phi_0 = \{u_0, v_0, w_0, h_0, d_0\}$ to preprocess the frames. Depending on how the $\Phi_0$ is obtained, there are two different evaluation protocols, frame based protocol and tracking protocol. The frame based protocol is the default for most of the current pose estimation work, in which the ground truth labels are used to obtain the $\Phi_0$. The tracking protocol only use the first frame ground truth to get the first frame initialization, and the following frames use the estimation of its previous frame to compute $\Phi_0$ in every following frame. The frame based protocol focus on evaluating the performance of the neural network in isolation, but the tracking protocol is more practical and requires the neural network to recover from inaccurate initialization. For both ground truth label induced and the estimation induced $\Phi_0$, we draw a bounding box around the 2D location of the joints to obtain $(u_0, v_0, w_0, h_0)$, and we compute the mean of the disparities (provided by the groundtruth or the previous frame estimation) as the $d_0$. For both initialization protocols, we use the average 3D estimation error in millimeters as evaluation metric. For both evaluation protocols, the initialization parameters in training process is always obtained from the ground truth. Figure 2 show the examples of the cropped images fed to the neural network.

We implement our system using TensorFlow. The network structure is based on the hourglass pose estimation framework. The 2D estimation network is a two stack hourglass network with intermediate supervision. We pick a point in the two stack hourglass network, and break the entire 2D location estimation network into two parts, $h_f$ and $h_{uv}$. Different break point provides different level of feature abstraction and speed tradeoff, which we further investigate in the ablation analysis section. The $h_d$ is one hourglass module on top of the concatenated feature maps. Both training stages are trained using RMSprop with batch size 32 for 100 epochs. The learning rate is set to 0.05 and decrease 70% for every 30 epochs. The network input size $W_n$ and $H_n$ are fixed as 256. The data augmentation strategies are determined by the evaluation protocol and training stage, as summarized in Table 1.

## 4.2 Ablation analysis

We first perform ablation study to evaluate the proposed neural network. In the proposed neural network architecture, the 2D estimator $h_f \circ h_{uv}$ is fixed as the two stack hourglass architecture, and the disparity estimation head is fixed as one hourglass module $h_{uv}$. We adopt these existing modules because of their consistently good performance on the 2D keypoint estimation tasks, but other more recent architectures, such as *CPM* [29] or the *SimpleBaseline* [30] can be used as well. However, evaluating the performance of these architectures is not our focus in this work.

|  | Left+Right | | | | Left Only | | | |
|---|---|---|---|---|---|---|---|---|
|  | D2S4 | D4S4 | D2S8 | D4S8 | D2S4 | D4S4 | D2S8 | D4S8 |
| (a) RSTB/Frame | **13.05** | 13.16 | 13.82 | 13.96 | 14.01 | 14.41 | 15.02 | 14.97 |
| (b) STB/Frame | 8.49 | **8.34** | 8.88 | 9.18 | 8.70 | 9.15 | 10.02 | 9.96 |
| (c) STB+RSTB/Frame | 7.71 | **7.18** | 8.68 | 8.67 | 8.84 | 8.60 | 9.58 | 8.94 |
| (d) STB/Track | **14.37** | 14.85 | 15.79 | 15.55 | 265.13 | 280.84 | 300.44 | 214.51 |
| (e) STB+RSTB/Track | 14.94 | **14.25** | 15.73 | 15.77 | 274.02 | 257.34 | 1000+ | 205.09 |
| Speed(FPS) | 1.06 | 1.05 | 1.28 | 1.29 | 1.09 | 1.08 | 1.32 | 1.31 |

Table 2: Ablation analysis results. The numbers (except the last row) are average 3D hand joint estimation errors in millimeters (the smaller the better). Each row reflects one set of training/eval protocol. *STB+RSTB* means pretrain on *RSTB*, finetune on *STB* and eval on *STB*. *Frame/Track* denote the two evaluation protocols. Different color stripes denote comparable numbers. (a) is the error measured the RSTB validation set, and (b)-(e) are errors measured on the STB validation set. *D*S** denote different network architecture variants.

As mentioned above, different *breakpoints $h_f$* in the 2D estimator reflects different level of feature abstraction and speed tradeoff. To understand this tradeoff, we compare the performance of two points, *D2* and *D4*. The structure of hourglass network is composed of the following layers: *input, convolution, residual block, maxpool, hourglass stacks*. Accordingly, we define *D2* as the output of the first convolution layer (the stride is 2), and define *D4* as the output of the first max pool layer (the stride is 4). In order to make the comparison more meaningful (*D2* is twice as larger as *D4*), we downsample them to the same size before feeding to the disparity estimation module. We apply two variations of the downsampling, *S4* and *S8*. *S4* downsample the feature maps to four times as smaller as the input image, *i.e.*, the stride is 4. *S8* downsample to stride 8. In summary, there are four network configurations, *D2S4*, *D4S4*, *D2S8*, and *D4S8* [3]. We also compare the performances of using the stereo pair (*Left+Right*) with those of using only the left image (*Left Only*). The average 3D hand joint estimation errors are listed in the Table 2. Given the datasets (STB/RSTB) and evaluation protocols (Frame/Track), the ablation study is further divided into a few train/eval configurations, which are shown as different rows in the Table 2. Note that the RSTB dataset renders each frame independently to maximize rendering efficiency, so there is no *Track* protocol on the RSTB dataset.

By comparing the performance of the *Left+Right* with *Left Only*, we find that the combing both views of the camera can significantly improve performance of the 3D hand pose estimation. While the *Left Only* configuration achieves decent performance on the *Frame* protocol, which is consistent with the recent works on 3D hand pose estimation from monocular RGB image, they completely fail on the *Track* protocol. In contrast, our two view configuration performs well in both protocols. This shows that the proposed network can capture the local features to figure out the sparse disparity and also recovers from initialization noises.

By comparing with and without RSTB pretrain (STB/Frame vs. STB+RSTB/Frame and STB/Track vs. STB+RSTB/Track), we find that the pretraining on the proposed synthetic data improves performance on the STB dataset. Specifically, for the *Frame* protocol, the best performance with RSTB pretrain (7.18) is better than that without the RSTB pretrain (8.34). Although the *Track* protocol relies largely on the network capability to recovers from initialization mistakes, for which the *RSTB* dataset (rendered independently) helps very little, pretraining still provides decent gain (14.25 vs. 14.37). Presumably because of the greater data variations on RSTB, the test errors on RSTB (13.05mm) is larger than those on the STB test set (7.18mm).

---

[3]Because stride of D4 is already 4, there is no downsampling in *D4S4*

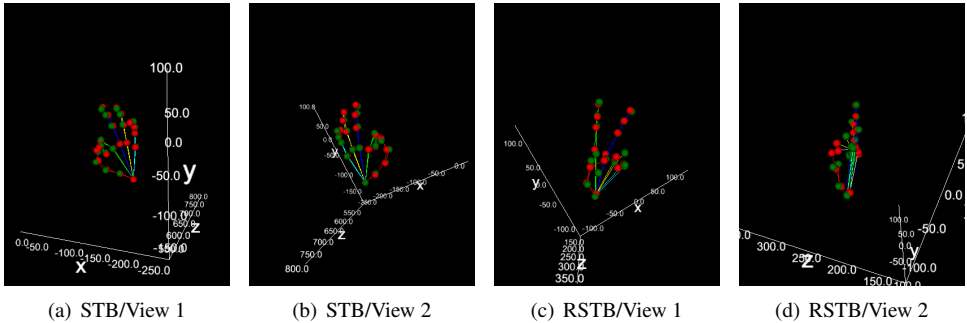

| (a) STB/View 1 | (b) STB/View 2 | (c) RSTB/View 1 | (d) RSTB/View 2 |

Figure 3: Result examples for the data sample in Figure 2. The red dots are the ground truth location, the green dots are the predicted location. Two views are selected for visualization.

The conclusion from comparing the *D4* and *D2* features is mixed, but the best performance is achieved using the *D4* features for the STB dataset on both the *Frame* and the *Track* protocols, which shows the benefits of using deeper features in the 2D estimation network to estimate disparity. Comparing the performance of *S4* and *S8*, we find that features with higher resolution perform better. In order to better understand the runtime of the architectures, in Table 2, we also measure the *frames per second (FPS)* statistics for these network variants. The statistics are collected using the default TF1.8.0, and on a CPU server with two cores and two threads. Averages of 20 runs after 5 runs of burn-in are reported. The runtime statistics shows that 1) the computation complexity addition from single view to two view networks is negligible, 2) while features with high resolution achieves better performance, it is also significantly slower. 3) using deeper features for disparity estimation does not add much computation, because the features are shared with the 2D estimation network, and the added computation is just for one of the views. Figure 3 shows qualitative results on the STB and RSTB dataset.

In order to show the effectiveness of the two-stage training mentioned in Sec. 3.3, we also conduct experiments using joint training. For the joint training, the losses $L_{uv}$ and $L_d$ are minimized together, *i.e.*, $h_f^*, h_{uv}^*, h_d^* = argmin_{h_f, h_{uv}, h_d}(L_{uv} + L_d)$. Because the disparity head and the 2D regressor are optimized together, certain data augmentation cannot be applied, as shown in Table 1. To reduce verbosity, we only report the results of joint training on the *D4S4* architecture, *Frame* protocol, and *STB* test set. The estimation errors for *STB/Frame* and *STB+RSTB/Frame* are 8.89mm and 8.80mm, respectively. Comparing with the estimation error of 8.34mm and 7.18mm in the case of two-stage training, we show that the two-stage training 1) reduces error rate for cases with and without synthetic data pretraining, 2) is particularly effective to leverage the synthetic training data.

## 4.3 Comparisons with the state of the art methods

There are two previous methods on 3D hand pose estimation using stereo cameras, one from Zhang *et al*. [34] and the other from Panteleris *et al*. [18]. Because of the code availability, we can only make comparison with the method proposed by Zhang *et al*. on the STB dataset. Using the high quality disparity map released by Zhang *et al*. [34], we evaluate the performance of HandPointNet [3], one of the state of the art methods for depth based 3D hand pose estimation, using the code released by the author. Another interesting baseline to compare with is to run the 2D keypoint estimator on the two views and compute the sparse disparity directly from the estimated 2D location of the joints. We denote this approach as *2D-Direct*. Because it is unknown how to run the *Track* protocol using the HandPointNet, we only compare these approaches on the *Frame* protocol. Also, because the code availability, we cannot obtain the high quality disparity map from the RSTB dataset using method of Zhang *et al*., and we can only train HandPointNet with only the STB dataset. The average 3D joint estimation errors are compared in the Table 3.

|  | Ours | Ours w/ RSTB | HandPointNet [9] | Direct-2D |
|---|---|---|---|---|
| average error (mm) | 8.34 | 7.18 | 7.6 | 24.60 |

Table 3: Compare with baselines on the STB dataset. *Ours* means the best as in Table 2.

Comparing the performance of *2D-Direct* and ours, we find that the intuitive two branch approach cannot estimate the disparity very well. This is partly because our disparity estimation module learns better features to estimate the disparity. Also note that the *2D-Direct* approach runs at only 0.69 FPS, much slower than ours (1.04 or 1.28), because it is required to run the heavy 2D estimation network on both views. Comparing with the depth based solution, our approach with RSTB pretraining outperforms the HandPointNet, and even without RSTB pretraining, our approach is still comparable. This shows that the data driven end-to-end approach is capable to outperform the depth based methods that explicitly encode the knowledge of stereo matching and other priors used in the method of Zhang *et al*. This is an encouraging result, because our approach is much more compact than the pipeline based on stereo matching and depth based inference.

Note that the proposed approach has potential limitations that require future study. 1) This approach cannot handle multiple hands or cases with hand/object interaction. 2) Demonstrating the proposed method actually out-speed existing works requires further study to benchmark more existing works and incorporate latest development on efficient deep learning, such as MobileNet [7] and Compression [6].

# 5    Conclusion

In this paper, we have proposed an end-to-end approach to estimate the full 3D hand pose from stereo cameras. We have developed a framework based on any 2D keypoint regressor to estimate the sparse disparity of the hand joints. To effectively learn the model, we have created a large scale synthetic dataset with stereo RGB images and full 3D hand pose annotations. In order to properly evaluate the full 3D hand pose estimation, we proposed a reference tracking algorithm and the companion evaluation protocol. While being more efficient and compact, our approach outperformed the existing approach based on the depth map from stereo matching. We have also shown that using stereo cameras can significantly improve the 3D hand pose estimation, especially in the tracking scenarios.

For future work, we plan to 1) solve the above mentioned limitations through collecting more diverse dataset, 2) optimize the neural network, especially the module to combine the left/right features, 3) extend the framework to other high level tasks, such as 3D human pose estimation, 3D reconstruction, 3D face alignment and recognition.